\documentclass[final]{l4dc2024}

\usepackage{enumerate}
\usepackage{enumitem}

\usepackage[font=small,labelfont=bf]{caption}

\title[Why Change Your Controller When You Can Change Your Planner]{Why Change Your Controller When You Can Change Your Planner: Drag-Aware Trajectory Generation for Quadrotor Systems}
\usepackage{times}

\coltauthor{\Name{Hanli Zhang}$^{\dagger}$ \Email{hanlizh@seas.upenn.edu}\\
 \Name{Anusha Srikanthan}$^{\dagger}$ \Email{sanusha@seas.upenn.edu}\\
 \Name{Spencer Folk} \Email{sfolk@seas.upenn.edu}\\
 \Name{Vijay Kumar} \Email{kumar@seas.upenn.edu}\\
 \Name{Nikolai Matni}\thanks{ This research is supported by NSF Grant CCR-2112665 and $^{\dagger}$these authors contributed equally to this work. \\ Link to code:  \url{https://github.com/hanlizhang/AeroWrenchPlanner}} \Email{nmatni@seas.upenn.edu}\\
 \addr University of Pennsylvania, Philadelphia, PA \\
 }

\begin{document}

\maketitle


\begin{abstract}%
  Motivated by the increasing use of quadrotors for payload delivery, we consider a joint trajectory generation and feedback control design problem for a quadrotor experiencing aerodynamic wrenches. Unmodeled aerodynamic drag forces from carried payloads can lead to catastrophic outcomes. Prior work model aerodynamic effects as residual dynamics or external disturbances in the control problem leading to a reactive policy that could be catastrophic. Moreover, redesigning controllers and tuning control gains on hardware platforms is a laborious effort. In this paper, we argue that adapting the trajectory generation component keeping the controller fixed can improve trajectory tracking for quadrotor systems experiencing drag forces. To achieve this, we formulate a \emph{drag-aware planning problem} by applying a suitable relaxation to an optimal quadrotor control problem, introducing a tracking cost function which measures the ability of a controller to follow a reference trajectory. This tracking cost function acts as a regularizer in trajectory generation and is learned from data obtained from simulation. Our experiments in both simulation and on the Crazyflie hardware platform show that changing the planner reduces tracking error by as much as 83\%. Evaluation on hardware demonstrates that our planned path, as opposed to a baseline, avoids controller saturation and catastrophic outcomes during aggressive maneuvers. 
\end{abstract}


\section{Introduction}
Quadrotor systems have found widespread use across various applications, such as agriculture~\citep{liu2018robust}, aerial photography~\citep{puttock2015aerial}, and delivery~\citep{brunner2019urban}. Their ease of use is due to collaborative research efforts addressing the challenges of modeling, estimation and quadrotor control as described by~\citet{hamandi2020survey, mahony2012multirotor}. Quadrotor control is known to be a difficult problem because it requires tracking smooth trajectories belonging to $SE(3)$. Prior work~\citep{mellinger2011minimum} solves quadrotor control by considering a tractable two-layer approach solving a reference trajectory generation problem at the \emph{top planning layer} using differential flatness (typically at a slower frequency). This reference trajectory is then sent to the \emph{low tracking layer} where a feedback controller~\citep{lee2010geometric}, operating in real time, attempts to follow the reference trajectory. With the advent of efficient planning algorithms, e.g.~\citep{csenbacslar2023probabilistic, LiuSE3SearchRAL18, mohta2018fast, gao2020teach, sun2021fast, herbert2017fastrack}, the two layer approach of trajectory planning and feedback tracking control has emerged as the gold standard. While conceptually appealing, the two layer approach has several shortcomings. A critical one is the lack of guarantees that the planned trajectories can be adequately tracked by the feedback controller. This could be due to various factors such as unmodeled dynamics, or saturation limits of the hardware in use for control. Some approaches propose novel ways to guarantee input feasibility~\citep{schweidel2023safe, mueller2015computationally} by assuming simplified dynamics models or disturbances. However, there are several tasks such as payload delivery that remain challenging for the quadrotor system~\citep{kumar2012opportunities}. 

For instance, transporting a package may introduce large external wrenches due to parasitic drag from the payload, which limits safe operation. These aerodynamic forces and torques are complex functions of variables such as relative airspeed, individual rotor speeds, attitude, rotation rate, and payload geometry, see~\citet{mahony2012multirotor} for a detailed characterization. Furthermore, such tasks introduce significant discrepancies between traditional rigid body dynamics models and the real world, resulting in large deviations of the system from planned trajectories. To circumvent effects of drag forces, several approaches in the literature~\citep{hoffmann2007starmac, huang2009aerodynamics, svacha2017improving, faessler2017differential, wu2022model} either learn a policy or change the controller design to account for unmodeled dynamics. Changing controller design is a reactive approach to aerodynamic effects since the controller has a myopic view of future states. This motivates the need for taking a proactive approach to preemptively account for deviations between reference and system trajectories. This leads to our next discussion on prior work that attribute errors as arising from differing model assumptions at each layer in the two-layer approach. 

To address challenges arising from inaccurate dynamics models, multirotor navigation systems \emph{impose} layered architectures in their decision making systems with coupled feedback loops~\citep{quan2020survey}. 
Prior work on multi-rate-control from~\citet{rosolia2020multi} propose a layered solution by planning using a reduced-order model-based MPC and control synthesis via control barrier functions for safety critical systems. 
Alternatively, there are also several papers~\citep{kaufmann2023champion, kaufmann2019beauty} that forego layered solutions, and instead opt for a black-box reinforcement learning based approach. 
Recent work by~\citet{srikanthan2023augmented, srikanthan2023data, matni2016theory} seek to derive layered architectures from an overall optimal control problem as a \emph{two-layer trajectory planner and feedback controller}, rather than imposing it as done previously in robotics applications. 
In this paper, we specialize contributions from~\citet{srikanthan2023data} for quadrotor control with aerodynamic wrenches and propose a novel solution to learning controller tracking cost without any dependence on discretization time of the feedback controller. 
Our contributions are as follows:
\begin{enumerate}[noitemsep,leftmargin=*]
    \item We formulate and solve a data-driven \emph{drag-aware trajectory generation} for a quadrotor system subject to aerodynamic wrenches via a relaxation of a quadrotor optimal control problem;
    \item We show it naturally yields a regularizer that captures the tracking cost of an $SE(3)$ feedback controller, and propose a data-driven approach to learning the tracking cost function;
    \item We design, demonstrate and open-source code and experiments that evaluate our method against a baseline~\citep{mellinger2011minimum} on an $SE(3)$ controller in simulation using RotorPy~\citep{folk2023rotorpy} and Crazyflie 2.0, showing significant improvements in position tracking error.
\end{enumerate}
In what follows, we discuss related work and present background work in Section~\ref{sec:related-work}. In Section~\ref{sec:methods}, we formulate and solve our proposed \emph{drag-aware trajectory generation} using a suitable relaxation of a quadrotor optimal control problem and show how it naturally leads to the inclusion of a regularizer that captures the tracking cost of an $SE(3)$ feedback controller. We then describe our supervised learning approach to learn a tracking cost function from trajectory data. In Section~\ref{sec:experiments}, we evaluate our proposed method and a baseline~\citep{mellinger2011minimum} in terms of position tracking error using RotorPy~\citep{folk2023rotorpy} as our simulator and demonstrate hardware experiments in Section~\ref{sec:hardware}. Finally, we discuss the impact and caveats of our proposed method and conclusion remarks in Section~\ref{sec:conclusion}.

\section{Related Work and Background}\label{sec:related-work}
We first discuss previous attempts at quadrotor control with aerodynamic wrenches and briefly highlight results from~\citep{srikanthan2023data} on data-driven dynamics-aware trajectory generation which accounts for tracking cost function of controllers in the planning layer. 

\subsection{Quadrotor control with aerodynamic wrenches}
Understanding and compensating for multirotor aerodynamic wrenches dates back to at least 2007 with the Stanford STARMAC II quadrotor aircraft~\citep{hoffmann2007starmac}. Here, the effects of blade flapping and parasitic drag were modeled and compared against experimental data. ~\citet{huang2009aerodynamics} applied these models to improving tracking control during aggressive stall turn maneuvers with impressive results. More recent work from \citet{svacha2017improving} demonstrated the use of modeling aerodynamic wrenches as a linear function of velocity as a way to improve the performance of an $SE(3)$ tracking controller. Linear models are favorable because they preserve the differential flatness property of quadrotors as described in~\cite{faessler2017differential}, but they come at the cost of model inaccuracy at higher speeds caused by parasitic drag on the airframe and any additional payloads. Other methods~\citep{antonelli2017adaptive, tal2020accurate} treat aerodynamic effects as disturbances and use disturbance rejection-based feedback controllers, or learn residual dynamics to model aerodynamic effects and design MPC for agile flight~\citep{torrente2021data}. Recently, quadrotor tracking control subject to aerodynamic wrenches has garnered the attention from deep learning communities, most notably culminating in works like Neural-fly~\citep{connell2022neural}. The authors train neural network control policies to compensate for aerodynamic wrenches based on as little as 12 minutes of flight data. However, note that all the approaches discussed so far directly modify the feedback control laws to account for parasitic drag or wind as opposed to changing the planned trajectories. Inspired by data-driven methods on dynamic feasibility of planners, we present dynamics-aware trajectory generation as an alternative to changing controller design.


\subsection{Layering as Optimal Control Decomposition}~\label{sec:layering-ocp}
Following~\citet{srikanthan2023data}, consider a finite-horizon, discrete-time nonlinear dynamical system
\begin{equation}\label{eq:dynamics}
x_{t+1} = f(x_t, u_t), \ t=0,\dots, N-1,
\end{equation}
with state $x_t \in \mathcal{X} \subseteq \mathbb{R}^n$, and control input $u_t \in \mathcal{U} \subseteq \mathbb{R}^k$ such that $\mathcal{X, U}$ are the domains of states and control inputs, respectively, at time $t$. The task is to solve the following constrained optimal control problem (OCP):
\begin{equation} \label{prob:master-problem}
\begin{array}{rl}
    \underset{x_{0:N},u_{0:N-1}}{\mathrm{minimize}} & \mathcal{C}(x_{0:N}) + \sum_{t=0}^{N-1}\|D_t u_t\|_2^2  \\
     \text{s.t.} & x_{t+1} = f(x_t,u_t), \, t=0,\dots,N-1,\\
     & x_{0:N} \in \mathcal{R},
\end{array}    
\end{equation}
where $\mathcal{C} : \mathcal{X}^{N+1} \rightarrow \mathbb{R}$ is a trajectory cost function with $\mathcal{X}^{N+1}:=\mathcal{X} \times \cdots \times \mathcal{X}$ (the Cartesian product of the feasible states), $D_0, D_1, ...., D_{N-1} \in \mathbb{R}^{k \times k}$
are matrices that penalize control effort, $\mathcal{R}$ defines the feasible region of $x_{0:N}$, and $x_{0:N}, u_{0:(N-1)}$ are the state and input trajectories over time horizon $N$. The OCP presented in~\eqref{prob:master-problem} is an essential component of control schemes for robotic applications where the trajectory cost function $\mathcal{C}$ captures high-level task objectives or reward smooth trajectories, whereas the state constraint $\mathcal{R}$ encodes obstacle avoidance, waypoint constraints, or other mission-specific requirements. 

As in~\cite{srikanthan2023data}, we first introduce a redundant reference trajectory variable $r_{0:N}$ (constructed by stacking the sequence of reference states $r_0, \cdots, r_N$) to the OCP~\eqref{prob:master-problem} and constrain it to equal the state trajectory, i.e., satisfying $x_{0:N}=r_{0:N}$ and relax this redundant equality constraint to a soft-constraint in the objective function
\begin{equation} \label{prob:relaxed-problem}
    \begin{array}{rl}
        \underset{r_{0:N}}{\mathrm{min.}}& \mathcal{C}(r_{0:N}) + \underset{x_{0:N}, u_{0:N-1}}{\mathrm{min.}} \sum_{t=0}^{N-1}\left(\lVert D_t u_{t} \rVert_2^2 + \rho \lVert r_t - x_t \rVert_2^2\right) + \rho \lVert r_N - x_N \rVert_2^2\\
        \text{s.t.}&r_{0:N} \in \mathcal{R}, \quad \quad \quad \text{ s.t. } x_{t+1} = f(x_t, u_t).
    \end{array}
\end{equation}
Problem~\eqref{prob:relaxed-problem} admits a layered interpretation: the inner minimization over state and input trajectories $x_{0:N}$ and $u_{0:N-1}$ is a traditional feedback control problem, seeking to optimally track the reference trajectory $r_{0:N}$. The outer optimization over the trajectory $r_{0:N}$ seeks to optimally ``plan'' a reference trajectory for the inner minimization to follow. Here the weight $\rho>0$ specifies the soft-penalty associated with the constraint $r_{0:N}=x_{0:N}$. The inner minimization defines a tracking penalty:
\begin{equation}\label{eq:tracking-cost}
\begin{aligned}
    g_{\rho}^{track}(x_0, r_{0:N}) :=
    \underset{x_{0:N}, u_{0:N-1}}{\mathrm{min}}&\quad \sum_{t=0}^{N-1}\left(\lVert D_t u_{t} \rVert_2^2 + \rho \lVert r_t - x_t \rVert_2^2\right) + \rho \lVert r_N - x_N \rVert_2^2\\
        \text{s.t.}&\quad \text{dynamics \eqref{eq:dynamics}}. 
\end{aligned}
\end{equation}
The tracking penalty $g_{\rho}^{track}(x_0, r_{0:N})$ captures how well a given trajectory $r_{0:N}$ can be tracked by a low layer control sequence $u_{0:N-1}$ given the initial condition $x_0$, and is naturally interpreted as the cost-to-go associated with an augmented system (see \S\ref{sec:learning}). Letting $\pi(x_t,r_{0:N})$ denote the feedback control policy which (approximately) solves problem~\eqref{eq:tracking-cost},~\citet{srikanthan2023data} use $g_{\rho,\pi}^{track}(x_0, r_{0:N})$ to denote the resulting cost-to-go that it induces. Whenever a closed-form expression for~\eqref{eq:tracking-cost} in terms of $r_{0:N}$ is difficult to compute analytically, it can be approximated using data-driven techniques as discussed in Section~\ref{sec:learning}.

Assuming that an accurate estimate of the tracking penalty~\eqref{eq:tracking-cost} can be obtained, the OCP~\eqref{prob:relaxed-problem} can now be reduced to the \emph{static} optimization problem (i.e., without any constraints enforcing the dynamics~\eqref{eq:dynamics}):
\begin{equation} \label{prob:layered-problem}
    \begin{aligned}
        \underset{r_{0:N}}{\mathrm{minimize}}&\quad \mathcal{C}(r_{0:N}) + g_{\rho, \pi}^{track}(x_0, r_{0:N}) \
        \text{s.t.} \, \quad r_{0:N} \in \mathcal{R}.
    \end{aligned}
\end{equation}
We view problem~\eqref{prob:layered-problem} as a family of trajectory optimization problems parameterized by $\rho$. For large $\rho$, optimal trajectories prioritize reference tracking performance while for small $\rho$, optimal trajectories prioritize minimizing the utility $\mathcal{C}$ cost. 

\subsection{Learning tracking penalty through policy evaluation}\label{sec:learning}
In~\citet[Sec. 4.2]{srikanthan2023data}, the authors define the following augmented dynamical system with states $\mu_t \in \mathbb{R}^{(N+1)n}$ and control inputs $u_t \in \mathbb{R}^{k}$. The state $\mu_t$ is constructed by concatenating the nominal state $x_t$ and the reference trajectory $r_{t:t+N}$ of length $N$ starting at time $t$, i.e., $\mu_t = (x_t, r_{t:t+N})\in \mathbb{R}^{(N+1)n}$. Letting $\mu_t^x := x_t$ and $\mu_t^r := r_{t:t+N}$, the augmented system dynamics can be written as $\mu_{t+1} = h(\mu_t, u_t)$, where
\begin{equation}\label{eq:aug_dyn}
    h(\mu_t, u_t) :=  \begin{bmatrix} f(\mu_t^x, u_t) \\ Z \mu_t^r\end{bmatrix}.
\end{equation}
Here $Z \in \{0,1\}^{Nn \times Nn}$ is the block-upshift operator, i.e., a block matrix with $I_n$ along the first block super-diagonal, and zero elsewhere. The state $\mu_t^x=x_t$ evolves in exactly 
the same way as in the true dynamics~\eqref{eq:dynamics}, whereas the reference trajectory $\mu_t^r := r_{t:t+N}$ is shifted forward in time via $Z\mu_t^r= r_{t+1:t+1+N}$.  Fixing policy $\pi(\mu_t)$, the policy dependent tracking cost becomes
\begin{equation}\label{eq:pi-tracking}
g_{\rho,\pi}^{track}(x_0, r_{0:N}) =
\sum_{t=0}^{N-1} \rho \left\lVert \mu_t^x - [\mu_t^r]_1 \right\rVert_2^2 + \lVert D_t u_t \rVert_2^2 + \rho \left\lVert \mu_N^x - [\mu_N^r]_1 \right\rVert_2^2.
\end{equation}
Finally, the proposed supervised learning approach approximates the tracking penalty \eqref{eq:tracking-cost} from data, refer~\citep{srikanthan2023data} for more details.

\vspace{-4pt}

\section{A Data-Driven Layered Decomposition for Quadrotor Systems}\label{sec:methods}
In this section, we describe how the methods outlined in Section~\ref{sec:related-work} can be applied to the quadrotor control problem subject to aerodynamic wrenches and derive our proposed~\emph{drag-aware trajectory generation} as an optimization problem. Finally, we discuss our proposed solution to learning the tracking penalty for an $SE(3)$ controller through policy evaluation. 
%
%
We consider the equations of motion of a quadrotor system subject to aerodynamic wrenches by adopting the dynamics presented in~\cite{folk2023rotorpy}:
\begin{equation}   \label{eq:quad_dynamics_drag}
\dot{x} := 
\begin{bmatrix}
\dot{p} \\
\dot{v} \\
\dot{R} \\ 
\end{bmatrix}
=
\begin{bmatrix}
v \\ 
\frac{1}{m}R\left(\begin{bmatrix} 0 \\ 0 \\ c_{f}\end{bmatrix} + f_{a}\right)-  g e_3\\ 
R [\omega]_\times \\ 
\end{bmatrix}
\end{equation}
where the state $x$ is composed of system position ($p \in R^3$), velocity ($v \in R^3$), and orientation ($R \in SO(3)$) with respect to the world frame, $[\cdot]_\times$ is the skew-symmetric matrix defined as $a\times b = [a]_\times b$, $c_f$ are the collective rotor thrusts, and $f_{a}$ are the aerodynamic forces given by: 
\begin{equation}
f_{a} = -C \vert \vert v \vert \vert_2 R^\top v - 
K \eta_{s} R^\top v
\end{equation}
Here, the matrices $C = \textbf{diag}(c_{Dx}, c_{Dy}, c_{Dz})$ and $K = \textbf{diag}(k_d, k_d, k_z)$ are parasitic and rotor drag coefficients,  respectively, and $\eta_s = \sum_{i=1}^4 \eta_i$ is the cumulative sum of the four individual rotor speeds, $\eta_i$. The drag coefficients are specified in the body frame axes, and the angular body rates are given by $\dot{\omega} = J^{-1}(\tau + m_a - \omega \times J \omega)$, where $J$ is the quadrotor's inertia matrix, $\tau = [\tau_x, \tau_y, \tau_z]^\top$ and $m_a$ are the moments arising from rotor thrusts and aerodynamics effects, respectively.

\subsection{Problem Formulation}\label{sec:problem-formulation}
Given a fixed feedback tracking controller such as an $SE(3)$ controller~\citep{lee2010geometric}, we pose the following question: \emph{how do we plan a dynamically feasible reference trajectory with low tracking error while running an $SE(3)$ controller in the presence of aerodynamic wrenches?} 

To formulate the problem, we consider a discrete-time dynamical model of the quadrotor system. From differential flatness~\citep{mellinger2011minimum}, we have that the dynamics of the quadrotor system derived from Euler-Lagrange equations is fully specified by the flat state $z_t \in \mathbb{R}^{17}$, consisting of its position $(p_t \in \mathbb{R}^3)$, velocity $(\dot{p}_t \in \mathbb{R}^3)$, acceleration $(\ddot{p}_t \in \mathbb{R}^3)$, jerk $(\dddot{p}_t \in \mathbb{R}^3)$, snap $(\ddddot{p}_t \in \mathbb{R}^3)$, yaw $(\psi_t \in \mathbb{R})$ and yaw rate $(\dot{\psi}_t \in \mathbb{R}^3)$. 
Discretizing the continuous-time quadrotor dynamics~\eqref{eq:quad_dynamics_drag} using explicit Runge Kutta method of order $5$, we define a discrete-time dynamical system evolving on the flat states as
\begin{equation}\label{eq:quad-discrete-time}
    z_{t+1}=f_q(z_t,u_t),\ t=0, 1, \cdots, N
\end{equation} 
Here, the subscript $q$ denotes that it is for a quadrotor system, $z_t$ are the flat states, 
and $u_t \in \mathbb{R}^4$, are the control inputs consisting of collective thrust from the four rotors and torque moments about the $x, y$ and $z$ axes in the body frame. Hence, it is sufficient to design a controller for the flat state $z_t$ without requiring roll and pitch angles (assumed to be stabilized by a low layer attitude controller).

Consider the OCP for the discrete-time dynamics~\eqref{eq:quad-discrete-time}:
\begin{equation} \label{prob:quad-master-problem}
\begin{array}{rl}
    \underset{z_{0:N},u_{0:N-1}}{\mathrm{minimize}} & \sum_{t=0}^N \|\ddddot p_t\|_2^2 + \| \dot \psi_t\|_2^2 + \sum_{t=0}^{N-1}\|u_t\|_2^2  \\
     \text{s.t.} & z_{t+1} = f_q(z_t,u_t), \, t=0,\dots,N-1,\\
     & z_{0:N} \in \mathcal{R},
\end{array}    
\end{equation}
where the trajectory cost function $\mathcal{C}(z_{0:N}) = \sum_{t=0}^N \|\ddddot p_t\|_2^2 + \| \dot \psi_t\|_2^2$, $D_t=I, \forall\ t$ and $\mathcal{R}$ specifies smoothness constraints on position, yaw and its higher order derivatives. The OCP~\eqref{prob:quad-master-problem} closely resembles the minimum snap trajectory generation from~\citep{mellinger2011minimum} except for nonlinear dynamics constraints and a controller effort term in the objective. OCPs of the form~\eqref{prob:quad-master-problem} are difficult to solve due to nonlinear dynamics constraints.



\subsection{Drag-aware trajectory generation}
Now, we introduce our \emph{drag-aware trajectory generation} for quadrotor systems with aerodynamic wrenches as a solution to solve OCP~\eqref{prob:quad-master-problem}. Applying the relaxation from Section~\ref{sec:layering-ocp} to OCP~\eqref{prob:quad-master-problem} for an $SE(3)$ geometric controller denoted by $\pi_q(z_t, r_t)$, we obtain the controller dependent tracking penalty $g_{\rho, \pi_q}^{track}(z_0, r_{0:N})$ from tracking the reference trajectory $r_{0:N}$. Framing the inner minimization in terms of $\Bar{\rho} = 1/\rho$, we obtain:
\begin{equation}\label{eq:quad-tracking-cost}
\begin{aligned}
    g_{\Bar{\rho}, \pi_q}^{track}(z_0, r_{0:N}) &:=
    \underset{x_{0:N}, u_{0:N-1}}{\mathrm{min}} \sum_{t=0}^{N-1}\left(\Bar{\rho} \lVert u_{t} \rVert_2^2 + \lVert r_t - z_t \rVert_2^2\right) + \lVert r_N - z_N \rVert_2^2\\
        \text{s.t.}&\quad x_{t+1} = f_q(z_t, u_t). 
\end{aligned}
\end{equation}
where the weight $\Bar{\rho} \geq 0$ specifies a penalty on controller effort\footnote{We use $\Bar{\rho}$ in the inner minimization instead of $\rho$ because $\Bar{\rho} \approx 0$ leads to stable training and naturally prioritizes improving tracking performance while large values of $\rho$ lead to unstable training regimes. Changing from $\rho$ to $\Bar{\rho}$ does not change the optimal controller but scales the optimal value of the tracking cost.} and when $\Bar{\rho} \approx 0$,  optimal trajectories prioritize the reference tracking performance.

If one were to apply supervised learning directly for the augmented system with state $\mu_t$ as defined in~\eqref{eq:aug_dyn}, the dimensionality of the input to the neural network will be of the order $17 N$ where $N=100$ for a time discretization of $0.01$ seconds. To remove dependence on any specific time discretization, we assume that the reference trajectories are polynomial splines of position and yaw ($p, \psi$) of fixed order $n$ for a given set of keyframes $m$. Rewriting $r_{0:N} = \mathcal{M}(t)\mathbf{c}$ and tracking penalty as $g_{\Bar{\rho}, \pi_q}^{track}(z_0, \mathbf{c})$, we specialize the layered problem~\eqref{prob:layered-problem} for quadrotor control to obtain the \emph{drag-aware trajectory optimization} problem as: 
\begin{equation}\label{prob:drag-aware-plan}
    \begin{array}{rl}
         \underset{\mathbf{c}}{\mathrm{minimize}}& \mathbf{c}^T H \mathbf{c} + g_{\Bar{\rho}, \pi_q}^{track}(z_0, \mathbf{c})  \
         \text{subject to}
         \, A \mathbf{c} = b
    \end{array}
\end{equation}
where $\mathbf{c} \in \mathbb{R}^{4mp}$ represents the stacked coefficients of $p, \psi$ polynomials across each segment, $H$ is the minimum snap cost represented in matrix form, $A, b$ define continuity and smoothness constraints at waypoints on position, yaw and it's higher order derivatives up to jerk.
\vspace{-4pt}

\subsection{Learning the controller cost through policy evaluation}
Solving for the tracking penalty~\eqref{eq:quad-tracking-cost} for quadrotor systems with aerodynamic wrenches running an $SE(3)$ controller is difficult. Hence, we adopt a supervised learning approach to approximately learn a function that maps coefficients to tracking cost. We use Monte Carlo sampling~\citep{sutton2018reinforcement} to generate a set of $\mathcal{T}$ trajectories, each over a time horizon $N_i$, given by 
\[ (z^{(i)}_{0:N_i}, u^{(i)}_{0:N_i-1}, r^{(i)}_{0:N_i}, \mathbf{c}^{(i)})_{i=1}^{|\mathcal{T}|}\] 
where $z_{0:N_i}^{(i)}$ and $u^{(i)}_{0:N_i-1}$ are the $i$-th state and input trajectories collected from applying $SE(3)$ feedback controller $\pi_q$ to track reference trajectories $r_{0:N_i}^{(i)}$ calculated from evaluating the polynomial coefficients $\mathbf{c}^{(i)}$.  We also compute the associated tracking cost labels $y^{(i)} := g_{\Bar{\rho},\pi_q}^{track}(z_0^{i}, \mathbf{c}^{(i)})$ and approximate the policy dependent tracking penalty~\eqref{eq:quad-tracking-cost} by solving the following supervised learning problem 
$$
\begin{array}{rl}
     \mathrm{minimize}_{g\in\mathcal G} \ \sum_{i=1}^\mathcal{T}(g(x_0^{(i)}, \mathbf{c}^{(i)}) - y^{(i)})^2,
\end{array}
$$
over a suitable function class $\mathcal{G}$ encompassing three-layer feedforward neural networks (MLPs). Finally, we solve~\eqref{prob:drag-aware-plan} using projected gradient descent. 
\vspace{-4pt}

\section{Simulation Experiments}\label{sec:experiments}
We run our simulation experiments and comparisons using RotorPy from~\cite{folk2023rotorpy} with default values of drag coefficients and consider two baseline methods: (i) ``minsnap" from~\cite{mellinger2011minimum} a state-of-the-art planner for $SE(3)$ tracking control, and (ii) ``minsnap+drag" from~\cite{svacha2017improving} a state-of-the-art method that directly changes controller design to account for drag, to show improvements in position tracking error. We use an implementation of~\citet{svacha2017improving} in RotorPy and do not tune control gains. 
We evaluate our method on $4$ different neural networks, trained on cost functions by varying $\Bar{\rho}$ over a set of values $ \{0, 0.1, 0.5, 1\}$ and compare the position tracking error of our approach against baseline methods.  
 
\textbf{Data Collection}\label{sec:data-collection}: In order to estimate the policy dependent tracking penalty~\eqref{eq:quad-tracking-cost} for $\Bar{\rho}=\{0, 0.1, 0.5, 1\}$, we solve a minimum snap trajectory generation problem given keyframe positions and yaw angles to collect data in the form of reference polynomial coefficients for $p, \psi$. We generate $200, 000$ trajectories by fixing the order of polynomial coefficients $n=7$ and total number of segments $m=3$. We sample each keyframe position from a uniform distribution over $[-10, 10]\: m$ while ensuring that each keyframe was at least $1 m$ and no further than $3 m$ apart. The yaw angle keyframes were randomly sampled between $[-\pi/2, \pi/2]\: rad$. This results in a total of $96$ coefficients per trajectory due to $3$ segments, $4$ variables and $8$ coefficients per segment. We roll out each trajectory on the closed-loop $SE(3)$ feedback tracking controller from the RotorPy simulator to get the cost labels as defined in~\eqref{eq:quad-tracking-cost} for each value of $\Bar{\rho}$. We note that the trajectories generated from our data collection approach leads to a spectrum of low, medium, and high tracking error cases. We also include cases where the minimum snap trajectory is infeasible, which in many cases caused the quadrotor system to crash and therefore accrue a high tracking cost.


 \begin{figure}[t]
    \centering
    \subfigure[Evaluation of minsnap vs our planner]
    {\label{fig:quad-traj}\includegraphics[width=\columnwidth, height=2.1in]{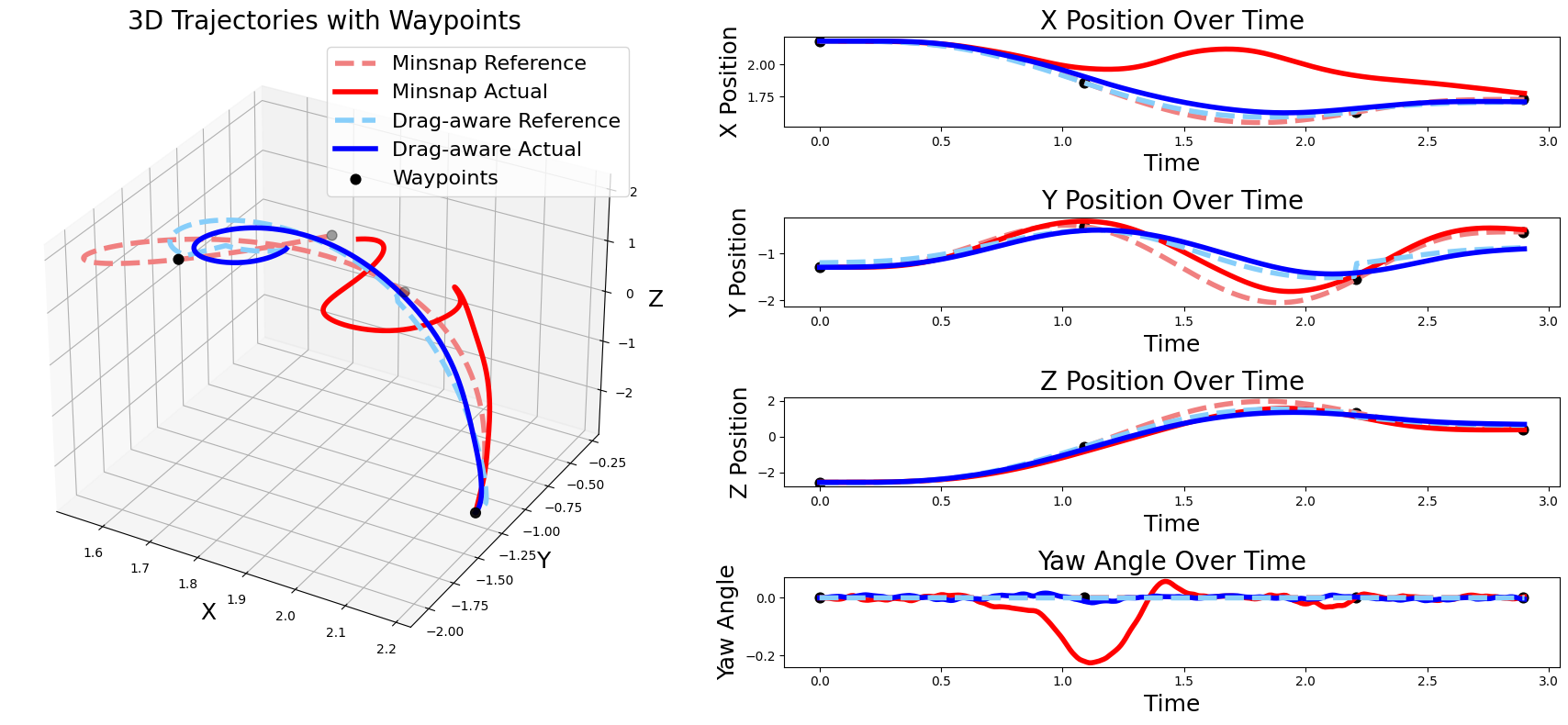}}
    \subfigure[Evaluation of minsnap+drag vs our planner]{\label{fig:drag-ctrl}\includegraphics[width=\textwidth, height=2.1in]{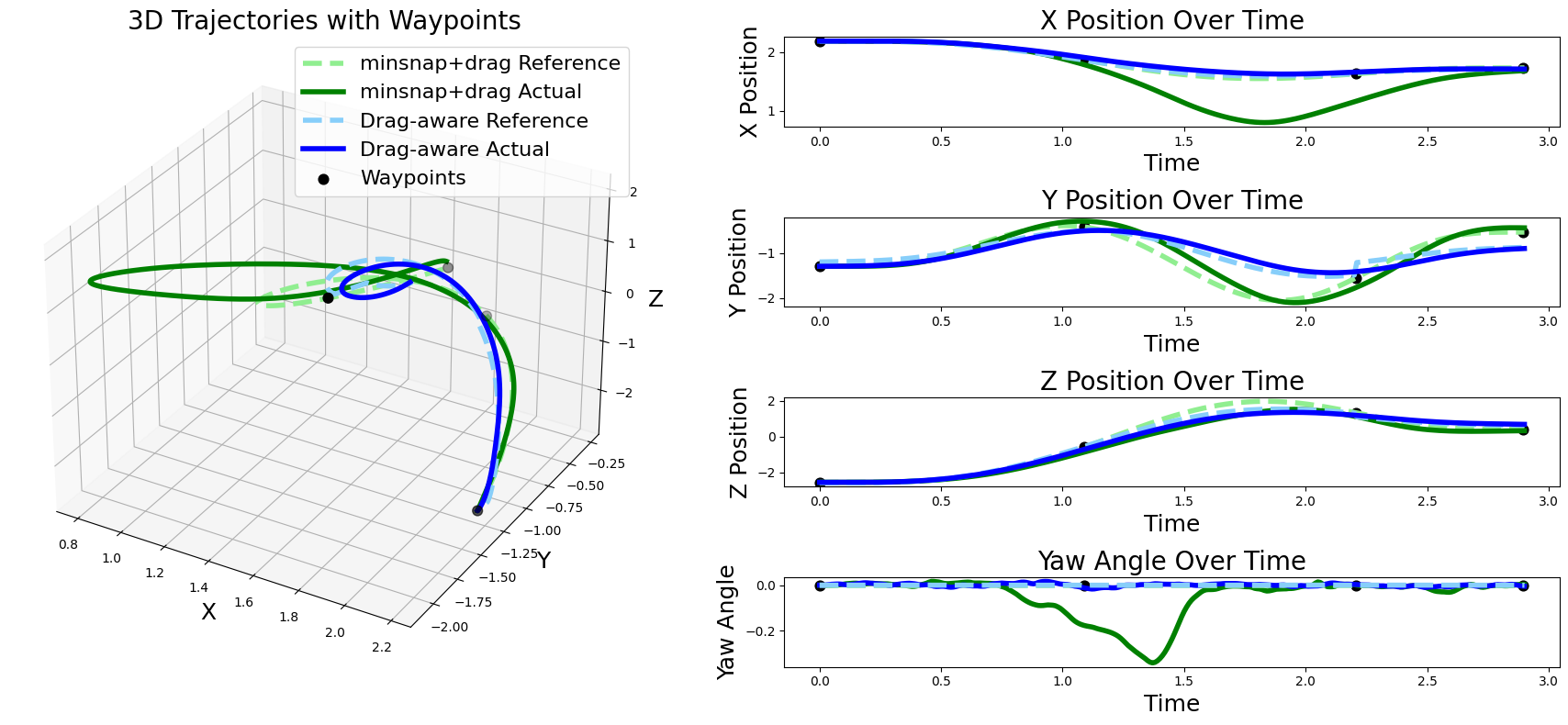}}
    \vspace{-6pt}
    \caption{We show a visualization of trajectories for $\Bar{\rho} = 0$ simulated on RotorPy where dotted and solid lines are reference and controller executed trajectories, respectively. On the left are 3D plots showing the deviation of controller executed trajectories and reference for our approach and baselines. On the right, we plot the $x, y, z$ and $\psi$ curves with waypoints to show the deviation in $x$ for baseline trajectories due to controller saturation while the drag-aware plan is tracked more accurately.}
    \vspace{-18 pt}
    \label{fig:3d-plots}
    \end{figure}
    
    \begin{figure}[t]
    \centering
    \subfigure[Tracking error vs time]
    {\label{fig:drag-aware-boxplot}\includegraphics[width=0.45\columnwidth, height=1.4in]{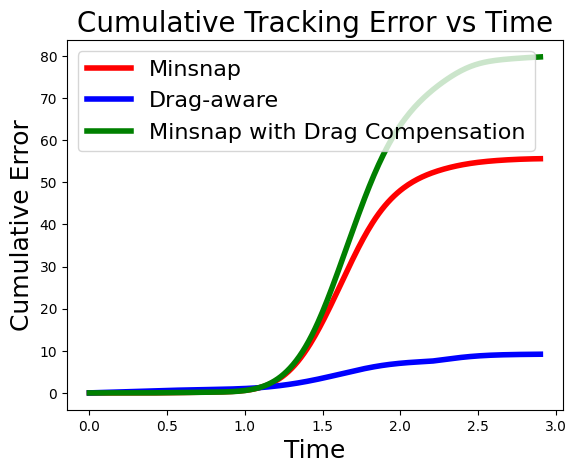}}
    \subfigure[Tracking error on a batch of trajectories]{\label{fig:error-vs-time}\includegraphics[width=0.45\textwidth, height=1.4in]{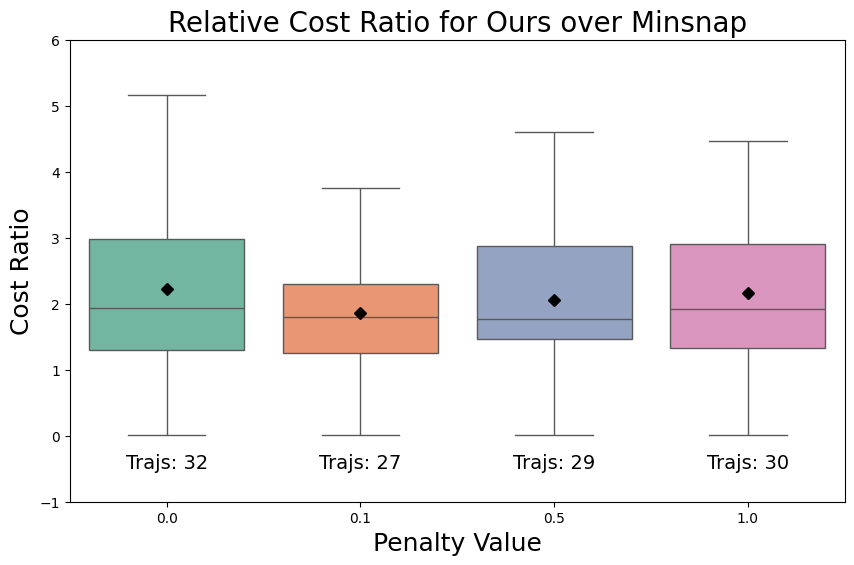}}
    \vspace{-6pt}
\caption{On the left, we plot cumumlative tracking error vs time for our approach and baselines for the trajectories visualized in Fig.~\ref{fig:3d-plots}. On the right, we plot the relative cost ratio of our approach over ``minsnap" baseline. The orange line represents the median ratio, separating the lower quartile from the upper quartile. The whiskers are the extreme values. We note that the tracking costs are obtained from the dynamics simulation (true cost) and not the network predictions.}
        \vspace{-18 pt}
    \label{fig:error-plots}
    \end{figure}


\textbf{Training}: We split the collected data into 80\% training and 20\% validation sets and train a multi-layer perceptron network with $3$ hidden layers of $\{100, 100, 20\}$ neurons, respectively, with Rectified Linear Unit (ReLU) activation functions.
We train a separate network for each value of $\Bar{\rho}$ using a batch size of $256$, learning rate $10^{-3}$ and run for $1000$ epochs. The entire network is set up using the optimized \texttt{JAX}, \texttt{Optax} and \texttt{Flax} libraries~\citep{jax2018github}. The loss function is optimized using stochastic gradient descent (SGD) with momentum set to $0.9$. At test time, i.e., when we compute trajectories to be tracked by the $SE(3)$ controller in RotorPy, we freeze the weights of the network and run projected gradient descent (PGD) using \texttt{jaxopt} to locally solve the drag-aware trajectory planning problem~\eqref{prob:drag-aware-plan} with maximum iterations set to $30$. 

\textbf{Results}: We evaluate the $SE(3)$ controller tracking penalty~\eqref{eq:quad-tracking-cost} on trajectories generated independently and in an identical way to the training dataset except that all yaw angles are set to $0$. We compare the tracking performance of our drag-aware planner with two baselines, ``minsnap" and ``minsnap+drag". Figure \ref{fig:3d-plots}a shows the full path of planned trajectories in dotted light coral and light sky blue, and controller executed trajectories in solid red and blue, from ``minsnap" and our approach, respectively. We observe that the trajectory planned by ``minsnap" results in controller saturation due to drag forces when trying to make a tight turn. In contrast, our drag-aware planner modifies the coefficients in such a way that the turn is feasible for the tracking controller while satisfying waypoint constraints. On the right, we plot the $p, \psi$ trajectories as a function of time. The system executed tracking of the ``minsnap" reference trajectory shown in red has large deviations in $x-$axis while our reference trajectory is more faithfully tracked by the system. In Figure~\ref{fig:3d-plots}b, we show the same trajectory executed on a modified $SE(3)$ controller with drag compensation without tuning gain constants. We found that, the controller with drag compensation fails to meaningfully modify the control inputs to avoid saturation. In Figure~\ref{fig:error-plots}a, we show the cumulative position tracking error over time of our approach and baselines. Our drag-aware planner improves the tracking error by around 83\% compared to baselines. As shown in Figure~\ref{fig:error-plots}b, we plot the relative ratio of tracking cost from ours vs ``minsnap" and observe that the tracking penalty in~\eqref{prob:drag-aware-plan} acts as a regularizer resulting in significant improvements in tracking cost when ``minsnap" cost is high and recovers a trajectory close to ``minsnap" when ``minsnap" cost is low. We note that the projected gradient descent solver from \texttt{jaxopt} doesn't always converge and in future work, we plan to solve convex approximations of the inference optimization for faster computation.  




\section{Hardware Experiments}\label{sec:hardware}

    \begin{figure}[t]
    \centering
    \subfigure[Minimum Snap Trajectory]
    {\label{fig:target parameter}\includegraphics[width=0.45\textwidth, height=1.4in]{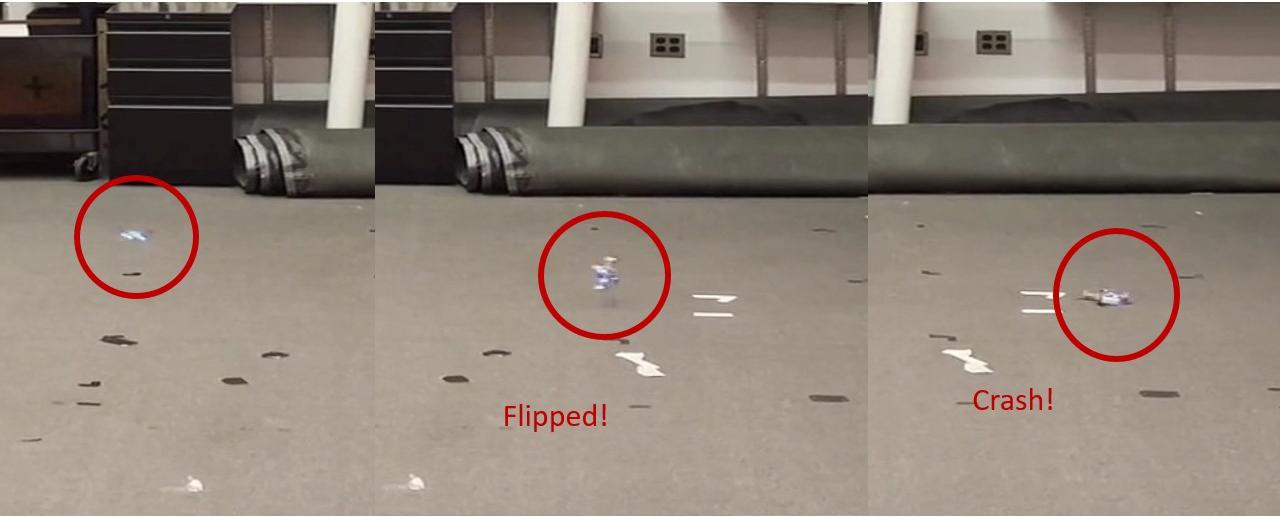}}
    \subfigure[Drag-aware Trajectory]{\label{fig:target stable}\includegraphics[width=0.45\textwidth, height=1.4in]{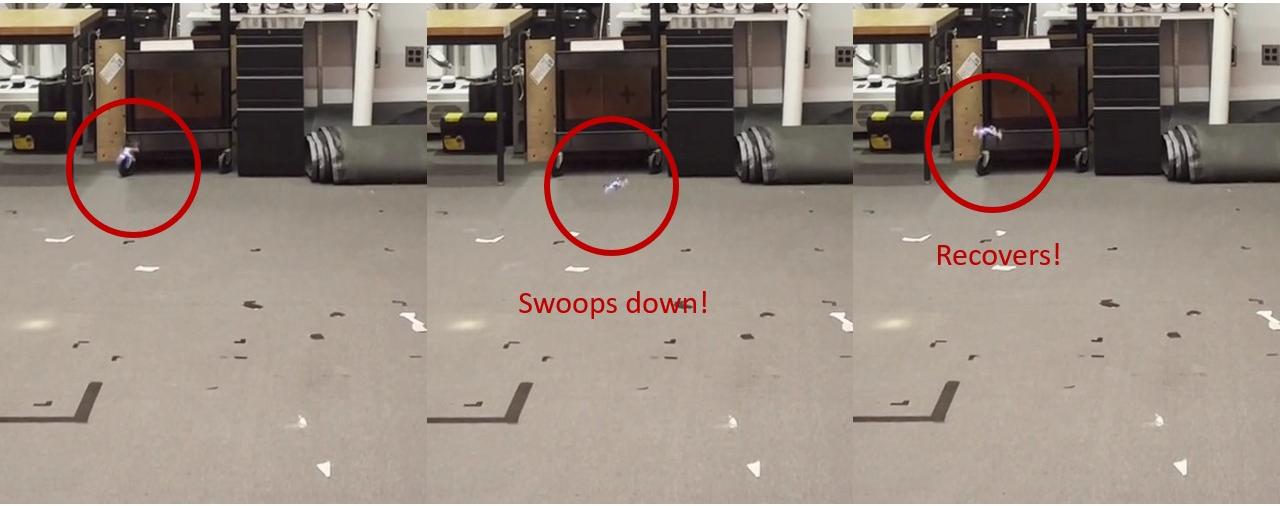}}
    \vspace{-6pt}
    \caption{We visualize snapshots of demonstrating trajectories planned by ``minsnap" as shown in a) and our approach as shown in b) on Crazyflie 2.0 and observe that the maneuver planned by the baseline results in controller saturation and crashing the quadrotor while our method planned a feasible path for the controller.}\vspace{-18 pt}
    \label{fig:hardware-demo}
    \end{figure}

We evaluate our trajectories on a Crazyflie 2.0\footnote{www.bitcraze.io} and use a motion capture system that provides measurements of pose and twist at 100Hz to a base station computer. On the base station, an $SE(3)$ controller that was tuned for the Crazyflie generates control commands in the form of a collective thrust and desired attitude. The Crazyflie uses onboard PID controllers and feedback from the inertial measurement unit (IMU) and motion capture system to track these control inputs.  

\textbf{Training and Results}: We generate a larger dataset ($709, 382$) of trajectories from RotorPy as described in Section~\ref{sec:data-collection} and trained a network with similarly chosen validation sets fixing $\Bar{\rho}=0$ and choosing a tracking cost function that uses mean position and velocity error. For evaluation, we select four waypoints within the motion capture space and plan a minimum snap trajectory using smooth polynomials of $7^{th}$ order for each segment. To evaluate our method, we then generate a modified reference trajectory solving the drag-aware planning problem~\eqref{prob:drag-aware-plan} using projected gradient descent on the same waypoints initializing the solver with the minimum snap trajectory. In both cases, time allocation is done by dividing the straight-path distance between each waypoint by an average velocity, which in our experiments was set to $2 m/s$. We visualize the hardware demonstration in Figure~\ref{fig:hardware-demo} showing snapshots of trajectories planned by ``minsnap" in a) resulting in controller saturation and crashing the quadrotor and our approach in b) successfully tracked by the controller. 

\section{Conclusion}\label{sec:conclusion}
We showed that the familiar two layer architecture composed of a trajectory planning layer and a low-layer tracking controller can be derived via a suitable relaxation of a quadrotor control problem for a quadrotor system with aerodynamic wrenches. The result of this relaxation is a \emph{drag-aware trajectory planning} problem, wherein the original state objective function is augmented with a tracking penalty which captures the closed-loop $SE(3)$ controller's ability to track a given reference trajectory.  We approximated the tracking penalty by using a supervised learning approach to learn a function from polynomial coefficients to cost from trajectory data.  We evaluated our method against two baselines on a quadrotor system  experiencing substantial drag forces, showing significant improvements of up to 83\% in position tracking performance. On the Crazyflie 2.0, we showed demonstrations where our method plans feasible trajectories while baseline trajectories lead to controller saturation, crashing the quadrotor. We conclude that a proactive approach to handling aerodynamic forces at the planning layer is a successful alternative to changing controller design. 
Future work will look to develop convex approximations of the proposed optimization for faster computation.

\bibliography{l4dc2024}

\end{document}